\documentclass[runningheads]{llncs}
\usepackage[T1]{fontenc}
\usepackage{graphicx}
\usepackage[utf8]{inputenc}
\usepackage{adjustbox}
\usepackage{lmodern}
\usepackage{microtype}
\usepackage{amsmath,amssymb,mathtools}
\usepackage{booktabs}
\usepackage{siunitx}
\usepackage{xcolor}
\usepackage{csquotes}
\usepackage{multirow}
\usepackage{array}
\usepackage{enumitem}
\usepackage{caption}
\usepackage{subcaption}
\usepackage[hidelinks]{hyperref}
\usepackage{color}
\usepackage[capitalize,nameinlink]{cleveref}
\usepackage{tikz}
\usetikzlibrary{arrows.meta,positioning,calc,fit}
\usepackage{svg}
\usepackage{float}

\title{Three-Step Hierarchical Transformer for Multi-Pedestrian Trajectory Prediction\thanks{This is the authors' preprint version. The final authenticated version will appear in the ICPR 2026 proceedings published by Springer.}}
\titlerunning{Three-Step Hierarchical Transformer}

\author{Raphaël Delécluse\inst{1,2,3}\and
Hazem Wannous\inst{1}\and
Laurent Grisoni\inst{2} \and
Laurent Guimas\inst{3}}
\authorrunning{R. Delécluse et al.}
\institute{IMT Nord Europe, University of Lille, CNRS UMR 9189 - CRIStAL, F-59000 Lille, France \and
University of Lille, CNRS UMR 9189 - CRIStAL, F-59000 Lille, France \and
Explain, F-59000 Lille, France}

\begin{document}
\maketitle

\begin{figure}[H]
    \centering
    \vspace{-0.8cm}
    \includegraphics[width=0.8\textwidth]{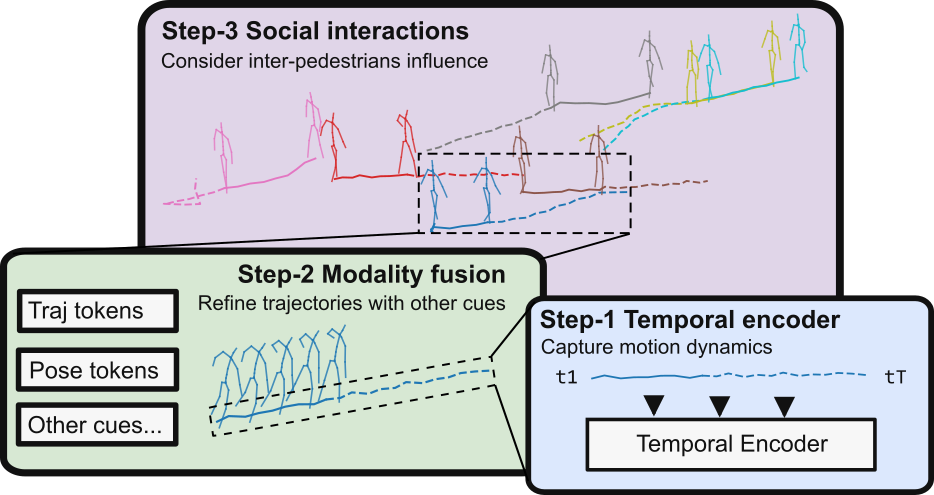}
    \caption{\emph{Progressive refinement in the Three-Step Transformer.} The model first predicts a coarse future trajectory from temporal tokens, then refines it through modality fusion and social attention to produce interaction-aware predictions.}
    \vspace{-0.6cm}
    \label{fig:abstract}
\end{figure}

\begin{abstract}

Pedestrian trajectory prediction requires modeling temporal dynamics, multimodal cues, and social interactions in crowded environments. Existing methods often address these factors separately or entangle them in costly attention blocks, limiting scalability, flexibility, and interpretability. We propose a three-step hierarchical Transformer that explicitly separates temporal encoding, multimodal fusion, and scene-level interaction reasoning. Lightweight GRU summaries enable efficient cross-modal attention, while social attention over time--agent tokens captures inter-pedestrian influences at manageable cost. Experiments on JTA, JRDB, and the Pedestrians and Cyclists in Road Traffic dataset show state-of-the-art performance on real-world datasets (JRDB, Urban) and competitive results on JTA. Ablation and qualitative analyses confirm the contribution of each stage and the model's ability to anticipate complex behaviors such as early turning. Code is available at \url{https://github.com/RaphaelDel/Three-step-hierarchical-transformer.git}.
\end{abstract}


\section{Introduction}\label{sec:intro}
Forecasting pedestrian trajectories is a fundamental problem in computer vision, robotics, and intelligent transportation systems, with applications ranging from autonomous navigation to crowd analytics and safety monitoring. The task is inherently challenging: human motion is influenced by multimodal cues such as body articulation, is shaped by complex social interactions, and unfolds within diverse and often cluttered environments. As a result, effective trajectory prediction requires simultaneously modeling temporal evolution, multimodal information, and interactions among multiple agents—while remaining computationally efficient enough to scale to real-world scenes.

Most existing approaches address only subsets of these challenges. Trajectory-only models capture local dynamics but struggle in scenarios where body pose or visual cues reveal early signs of intent. Multimodal models exploit richer cues but often fuse them through simplistic concatenation or modality-specific backbones, limiting flexibility and scalability. Social forecasting methods introduce inter-agent reasoning, yet typically merge temporal, multimodal, and social signals within a single attention block, making the model harder to interpret and computationally expensive. These limitations motivate the need for a unified framework that can handle heterogeneous modalities, variable numbers of pedestrians, and long temporal sequences in a principled and efficient manner.

In this work, we introduce a three-step hierarchical Transformer architecture that decomposes trajectory forecasting into complementary reasoning levels. A temporal encoder processes each modality independently, capturing short- and long-term motion dynamics using modality-specific GRUs or Transformers. A modality decoder integrates the resulting embeddings through cross-modal attention, enabling the target pedestrian’s trajectory to be refined using pose, bounding boxes, or other cues. Finally, a scene-level Transformer models inter-pedestrian interactions over time–agent tokens, providing social context without incurring the quadratic cost of attending jointly across time, modalities, and agents. This structured factorization makes the architecture interpretable, modular, and computationally efficient.\\


Our contributions are threefold:
\begin{itemize}
    \item We propose a novel three-step hierarchical Transformer that decouples temporal, multimodal, and social reasoning, offering a flexible and interpretable architecture for multi-pedestrian forecasting.
    \item We introduce a computationally and memory-efficient multimodal fusion strategy based on GRU summarization and decoder-style cross-modal attention, enabling scalable integration of heterogeneous cues.
    \item We provide extensive quantitative, qualitative, and ablation results demonstrating state-of-the-art performance on real-world datasets and competitive on  synthetic one, along with detailed analyses of success and failure modes.
\end{itemize}

Together, these contributions demonstrate that structured, hierarchical modeling provides a powerful and efficient foundation for future trajectory prediction.

\section{Related Work}\label{sec:related}
\subsection{Sequence Modeling for Trajectory Prediction}

Human trajectory forecasting is commonly formulated as a sequence-to-sequence prediction problem, where future motion is inferred from a history of observed positions. 
Early approaches such as Social-LSTM~\cite{alahi_social_2016} and SR-LSTM~\cite{zhang_sr-lstm_2019} modeled individual motion dynamics using recurrent neural networks (RNNs) while aggregating neighborhood information through social pooling. 
Subsequent work introduced graph-based representations~\cite{kosaraju_social-bigat_2019,mohamed_social-stgcnn_2020,yu_spatio-temporal_2020} to better encode structured interactions, and attention mechanisms~\cite{vemula_social_2018,giuliari_transformer_2021} to learn context-dependent relationships among pedestrians. 
Although these RNN and graph-based methods achieved good performance, they are limited in capturing long-range temporal dependencies and global context. 

Transformer architectures~\cite{vaswani_attention_2017} have since become the foundation for sequence modeling in this field. 
Their self-attention mechanism allows for parallel processing and improved modeling of both short and long-term temporal dependencies. 
Works such as AgentFormer~\cite{yuan_agentformer_2021} and ST-Transformer~\cite{yu_spatio-temporal_2020} have shown that attention-based reasoning can effectively represent temporal and relational dependencies across multiple agents. 
Further improvements include hierarchical temporal modeling~\cite{girgis2021latent,saadatnejad_social-transmotion_2023}, spatial-temporal disentanglement~\cite{liang_stglow_2024}, and hybrid recurrent-transformer structures that balance efficiency and temporal expressiveness. 
In a complementary direction, Zaier et al.~\cite{Zaier:WACV2025} introduce a compact manifold-valued representation for 3D human skeleton motion within a transformer-based framework. 
Their approach maps motion sequences onto a geometric manifold to encourage smooth and coherent long-term predictions, while combining Kendall's shape space to model non-rigid deformations with Lie group representations to capture rigid transformations.
Inspired by these trends, our temporal encoder combines GRU and Transformer mechanisms depending on the modality, allowing scalable yet flexible temporal representation learning.

\subsection{Multimodal and Social Reasoning}

Beyond purely positional inputs, recent works have emphasized the integration of multiple cues—such as human pose, bounding boxes, and visual semantics—to enrich motion prediction. 
Social-Pose~\cite{gao_social-pose_2025} demonstrated that 3D body articulation provides strong intent cues that improve trajectory forecasting. 
Social-TransMotion~\cite{saadatnejad_social-transmotion_2023} unified multi-agent and multi-modality reasoning within a Transformer-based framework, enabling explicit modeling of pose, trajectory, and social context through shared attention layers. 
Similarly, Cross-Feat~\cite{marchetti_crossfeat_2024} proposed a GRU-based feature compression scheme to manage modality dimensionality, and Cross-Modal Attention~\cite{zaier_cross-modal_2023} leveraged cross-attention between trajectory and pose embeddings to improve multimodal fusion. 
Our work builds upon these ideas but introduces a hierarchical separation between temporal encoding, modality decoding, and social reasoning, allowing interpretable and scalable fusion of heterogeneous signals.

Social interaction modeling has also evolved from handcrafted social-force models~\cite{helbing_social_1995} to learned relational reasoning. 
Graph-based networks~\cite{kosaraju_social-bigat_2019,mohamed_social-stgcnn_2020,yu_spatio-temporal_2020} and Transformer attention~\cite{yuan_agentformer_2021,saadatnejad_social-transmotion_2023} are now standard approaches for representing agent-to-agent influence without manual interaction rules. 
However, most existing models merge social and modal attention into a single reasoning block, which can blur the distinction between intra-agent (modal) and inter-agent (social) dependencies. 
In contrast, our design decouples these processes through a dedicated modality decoder followed by a scene transformer that performs cross-agent attention over time–agent tokens, ensuring clear interpretability and modular scalability.

Finally, several works have explored the inclusion of environmental and contextual cues to better understand scene semantics. 
Robot That Can See~\cite{salzmann_robots_2023} incorporated visual context through image-based encoders, while EmLoco~\cite{taketsugu_physical_2025} learned global scene embeddings directly from trajectories to enhance spatial awareness. 
These studies highlight the importance of contextual reasoning in motion forecasting. 
Our scene transformer follows a similar intuition but learns such context implicitly through attention across pedestrians, avoiding explicit scene-map supervision while maintaining the ability to reason about spatial configurations.

\section{Method}\label{sec:method}
\subsection{Problem Formulation}

We consider the problem of predicting the future trajectory of pedestrians in a scene, given observations of multiple agents and multiple sensory or derived modalities over a fixed temporal window.

Let $T$ denote the number of observed frames, $N$ the number of pedestrians in the scene, and $M$ the number of available modalities (e.g., 2D trajectory, 3D pose, bounding box, velocity, or other geometric descriptors). Each pedestrian $i \in \{1,\dots,N\}$ is represented at each time step $t \in \{1,\dots,T\}$ by a set of modality-specific features:
\[
\mathcal{X} = \left\{ \mathbf{x}^{(i,m)}_t \in \mathbb{R}^{d_m} \;\middle|\; i=1..N,\; m=1..M,\; t=1..T \right\},
\]
where $d_m$ denotes the dimensionality of modality $m$.

The objective is to predict the future trajectory of one or more target pedestrians over a horizon of $\Delta$ frames:
\(
\hat{\mathbf{y}}^{(i)}_{T+1:T+\Delta} = f_\theta(\mathcal{X}),
\)
where $f_\theta$ is a learnable model parameterized by $\theta$.

Unlike trajectory-only approaches, this formulation explicitly incorporates heterogeneous modalities and interactions between multiple agents. Both the number of pedestrians $N$ and the number of modalities $M$ may vary across scenes, and some modalities may be partially missing.

This variability makes flat representations inefficient, as dependencies across time, modalities, and agents are structurally different. The dimensionality of the problem scales with $T \times N \times M$, while interactions across these axes are not uniform. We therefore adopt a hierarchical design that decomposes reasoning into three complementary levels: (1) temporal modeling within each trajectory, (2) cross-modal fusion within each pedestrian, and (3) inter-pedestrian reasoning at the scene level. This structured decomposition forms the foundation of the architecture described in Section~\ref{sec:architecture}.

\subsection{Overall Architecture}
\label{sec:architecture}

The proposed architecture follows a three-step hierarchical design that explicitly separates temporal modeling, multimodal conditioning, and social interaction reasoning. Each level operates on a different structural axis of the data, allowing the model to scale efficiently with respect to the number of frames, modalities, and pedestrians while preserving interpretability.

The first level, the Temporal Encoder, processes pedestrian trajectories independently across time. It captures motion dynamics and extrapolates observed trajectories into an initial future hypothesis over the horizon $T+\Delta$, without considering other modalities or interactions. This stage focuses exclusively on temporal consistency and short-term forecasting.

The second level, the Modality Decoder, refines these trajectory hypotheses by conditioning them on auxiliary modalities such as pose or bounding-box information. A decoder-style cross-attention mechanism integrates modality-specific embeddings into the trajectory stream at each time step, enabling multimodal reasoning while remaining robust to missing or variable modality sets.

The third level, the Scene Transformer, operates on the predicted trajectories of all pedestrians jointly. By attending over the flattened set of time–pedestrian tokens, it models future social interactions and adjusts individual trajectories accordingly. This design allows the network to reason about interactions that may occur during the prediction horizon, rather than relying solely on past observations.

Finally, a lightweight Prediction Head composed of a multi-layer perceptron (MLP) maps the socially refined embeddings of the target pedestrian to the final predicted trajectory. Together, these components form a unified hierarchical framework that progressively refines motion predictions through temporal, multimodal, and social reasoning.

\begin{figure}[t]
    \centering
    \includegraphics[width=1.0\textwidth]{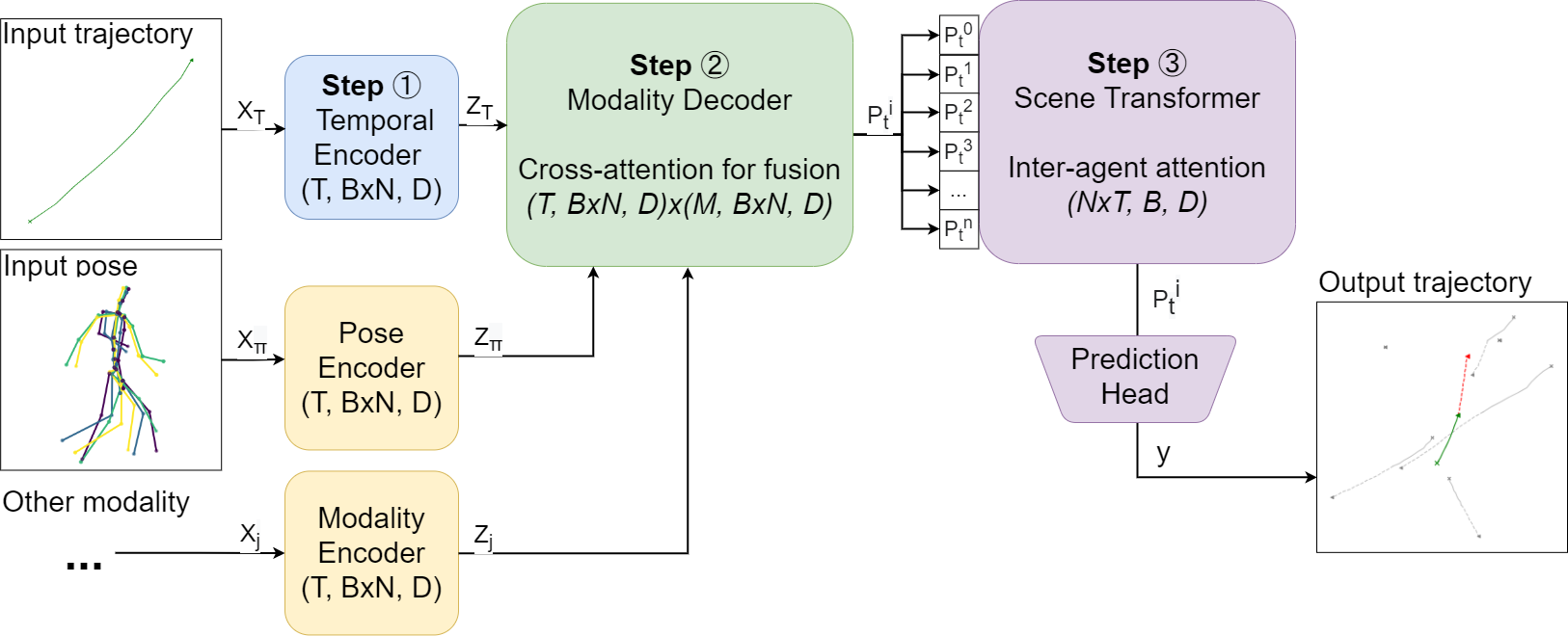}
    \caption{\emph{Overview of the proposed three-step hierarchical architecture.}
    (1) A temporal encoder extrapolates pedestrian trajectories from past observations into an initial future hypothesis.
    (2) A modality decoder refines this hypothesis by conditioning trajectory embeddings on auxiliary modalities via cross-attention.
    (3) A scene transformer jointly reasons over predicted trajectories to model social interactions and produce the final forecast.}
    \label{fig:architecture}
\end{figure}

\begin{figure}[t]
    \centering
    \includegraphics[width=0.7\textwidth]{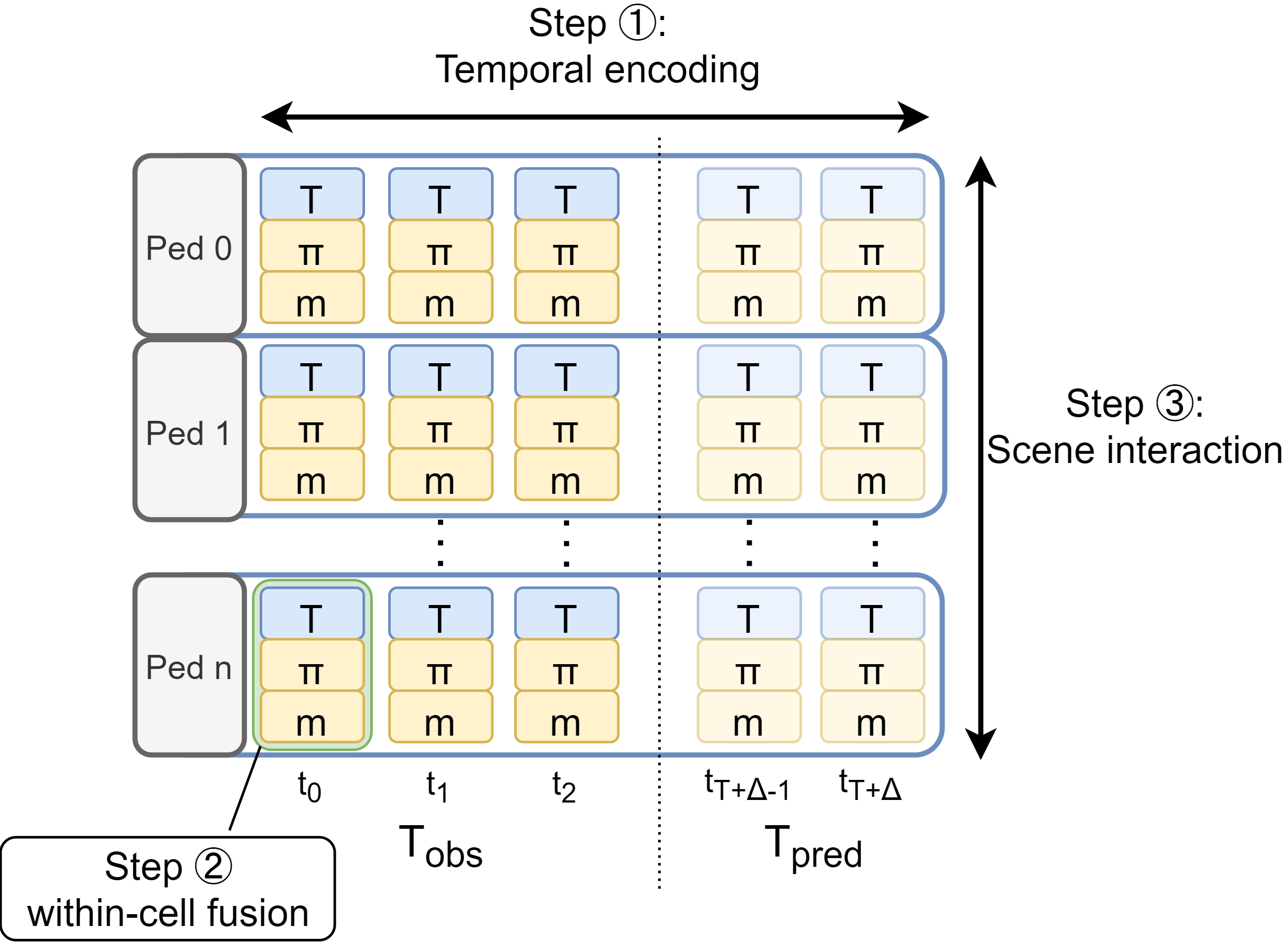}
    \caption{Overview of the three-step hierarchical architecture. Each (t,n) cell contains M tokens ; trajectory: T, pose: $\pi$ and m: another modality, embedded $\in \mathbb{R}^d$. \textit{Step 1} attends over time t per pedestrian ; \textit{Step 2} fuses T and $\pi$ by cross-attention ; \textit{Step 3 }attends over the flattened TxN token sequences with agent-aware encodings to preserve identity.}
    \label{fig:token_archi}
\end{figure}

\subsection{Temporal Encoder}

The temporal encoder constitutes the first level of the architecture and focuses exclusively on modeling the temporal evolution of pedestrian trajectories (Step 1 in Fig.~\ref{fig:architecture} and Fig.~\ref{fig:token_archi}). 
At this stage, no cross-modal or inter-agent reasoning is performed: each pedestrian trajectory is processed independently to capture motion dynamics such as velocity trends, directional changes, and temporal continuity.

\paragraph{Input representation.}
Given a batch of observed trajectories, the input tensor has shape $(B, T, N, K)$, where $B$ is the batch size, $T$ the number of observed frames, $N$ the number of pedestrians, and $K$ the trajectory feature dimension (e.g., 2D or 3D coordinates). 
A shared MLP first projects trajectory features into a common latent space of dimension $D$, yielding embeddings of shape $(B, T, N, D)$.
At this level, pedestrians are treated independently, and the effective input to the temporal encoder is reshaped as $(T, B \times N, D)$.

\paragraph{Composite positional encoding.}
To encode both temporal order and pedestrian identity, we apply a composite positional encoding to the trajectory embeddings.
The embedding dimension is split into two halves: sinusoidal temporal positional encodings are added to the even dimensions, while learned pedestrian identity embeddings are added to the odd dimensions.
This design allows the model to disentangle global temporal progression from agent-specific motion patterns while preserving identity information throughout the sequence.

\paragraph{Temporal forecasting.}

The temporal encoder is implemented as a Transformer operating along the time dimension. To produce a representation of length $T+\Delta$, we append $\Delta$ learnable \emph{future query} tokens to the $T$ observed trajectory tokens. A causal attention mask allows each future token to attend to all observed tokens (and optionally to earlier future tokens), enabling the model to extrapolate the observed motion into the future in a single forward pass. The resulting sequence of embeddings $\{z_t\}_{t=1}^{T+\Delta}$ serves as an initial trajectory hypothesis over the full horizon, which is further refined by multimodal fusion and social interaction reasoning in subsequent stages.

\subsection{Modality Decoder}

The modality decoder forms the second level of the architecture and integrates heterogeneous information sources for each pedestrian (Step 2 in Fig.~\ref{fig:architecture} and Fig.~\ref{fig:token_archi}). 
Its role is to refine the temporally encoded trajectory by conditioning it on auxiliary modalities such as pose, bounding boxes, or kinematic descriptors.

\paragraph{Modality embeddings.}
Each modality is first projected into the common latent space $\mathbb{R}^D$ using a modality-specific MLP.
After projection, all modalities are represented as sequences of shape $(T, B \times N, D)$.
To limit computational complexity, each modality sequence is summarized into a single embedding per pedestrian using a lightweight temporal encoder (e.g., a GRU), producing a tensor
$\mathbf{Z}_{\text{mod}} \in \mathbb{R}^{M \times (B \times N) \times D}$,
where $M$ denotes the number of available modalities.
This summary captures modality-specific temporal cues while avoiding attention over full temporal sequences.

\paragraph{Cross-modal decoding.}
The modality decoder follows a Transformer decoder architecture.
Trajectory embeddings of shape $(T, B \times N, D)$ act as queries, while the modality embeddings serve as keys and values:
\[
\hat{\mathbf{Z}}_{\text{traj}} = \text{Decoder}(\mathbf{Z}_{\text{traj}}, \mathbf{Z}_{\text{mod}}).
\]
Through cross-attention, the decoder injects complementary semantic and kinematic information into the trajectory stream, refining motion predictions using pose and other visual cues.
The impact of this multimodal conditioning is quantitatively analyzed in the ablation study presented in Section~\ref{sec:ablations}.
Missing modalities are handled naturally through attention masking, allowing the model to operate with variable modality sets.

This decoder-level fusion enables effective multimodal reasoning while keeping computational cost linear in the number of modalities.

\subsection{Scene Transformer}

The scene transformer constitutes the third and final level of the architecture and models inter-pedestrian interactions at the scene level (Step 3 in Fig.~\ref{fig:architecture} and Fig.~\ref{fig:token_archi}). 
At this stage, each pedestrian is already associated with a complete trajectory hypothesis over the horizon $T+\Delta$, obtained through temporal extrapolation and multimodal conditioning.
The role of the scene transformer is therefore not to generate motion from scratch, but to refine these predictions by accounting for social interactions that may occur between agents.

\paragraph{Input representation.}
After modality decoding, each pedestrian $i$ is represented by a sequence of embeddings 
$\{\mathbf{p}^{(i)}_t \in \mathbb{R}^D\}_{t=1}^{T+\Delta}$ 
that encode trajectory dynamics and multimodal cues.
To enable joint reasoning over time and agents, we flatten the temporal and pedestrian dimensions and form a set of tokens
\[
\mathbf{P} \in \mathbb{R}^{(N \times (T+\Delta)) \times B \times D},
\]
This representation exposes potential interactions not only at individual frames, but across the entire predicted motion of the scene.

\paragraph{Social interaction refinement.}
A Transformer encoder is applied to the flattened token set to perform self-attention across pedestrians and time:
\[
\tilde{\mathbf{P}} = \text{Transformer}_{\text{scene}}(\mathbf{P}).
\]
Because each token already encodes a future motion hypothesis, attention allows the model to reason about forthcoming social events such as collisions, following behaviors, or group formation, and to adjust trajectories accordingly.
Absent pedestrians are handled through masking, enabling variable crowd sizes.

\paragraph{Output and prediction.}
The refined embeddings are reshaped back into per-pedestrian sequences 
$\{\tilde{\mathbf{p}}^{(i)}_t\}_{t=1}^{T+\Delta}$.
For the pedestrian of interest, the embeddings corresponding to the future horizon are passed to the prediction head (MLP) to produce the final trajectory.
By operating on predicted trajectories rather than raw observations, the scene transformer acts as a dedicated social reasoning module that improves consistency and plausibility in multi-pedestrian settings.

\subsection{Computational and Memory Efficiency of the Three-Step Transformer}
\label{sec:efficiency}

Let $T$ be the observation length, $\Delta$ the prediction horizon, $N$ the number of pedestrians, and $M$ the number of available modalities (trajectory, pose, \emph{etc.}). A naïve transformer would flatten all tokens jointly (time $\times$ modality $\times$ pedestrian), yielding a sequence length
$L_{\text{naive}}=(T+\Delta)MN$ and quadratic attention memory $\mathcal{O}(L_{\text{naive}}^2)$.

Our model reduces this cost by factorizing attention into three stages aligned with the structure of the data.
\textbf{Step~1} performs temporal encoding independently per pedestrian, attending only along time.
\textbf{Step~2} performs cross-modal decoding: each auxiliary modality is first projected to the common latent space and summarized into a single token per pedestrian (e.g., via a lightweight temporal encoder such as a GRU), producing $M$ modality tokens per pedestrian. Trajectory tokens act as queries, while modality tokens provide keys/values.
\textbf{Step~3} performs scene-level social reasoning by attending over the flattened time--pedestrian set.

With $L_1=T+\Delta$ and $L_3=(T+\Delta)N$, the resulting attention-memory scaling is:
\begin{align}
\textbf{Step 1 (Temporal)}\!:~ &\mathcal{O}\!\left(N\,L_1^{2}\right), \nonumber\\
\textbf{Step 2 (Cross-modal)}\!:~ &\mathcal{O}\!\left(N\,L_1M\right), \nonumber\\
\textbf{Step 3 (Scene)}\!:~ &\mathcal{O}\!\left(L_3^{2}\right), \qquad
\Rightarrow\quad
\mathcal{O}\!\left(N\,L_1^{2}+N\,L_1M+L_3^{2}\right). \nonumber
\end{align}

Compared to $\mathcal{O}\!\left(((T+\Delta)MN)^2\right)$, this is typically much smaller when $M$ and $N$ are greater than a few units. Importantly, the three stages do not correspond to a simple partition of a single token sequence; each stage operates on a compressed representation tailored to its specific role. This design makes the full architecture computationally scalable while preserving the expressiveness required for multi-pedestrian trajectory forecasting.

\section{Experiments}\label{sec:experiments}
This section presents the experimental setup used to evaluate our hierarchical architecture for multi-pedestrian trajectory prediction. 
We begin by describing the datasets, evaluation metrics, and implementation details. 
We then report quantitative and qualitative results, showing that the proposed model achieves state-of-the-art performance across synthetic and real environments. 
Finally, we conduct a detailed ablation study analyzing the contribution of each transformer level—temporal encoding, multimodal fusion, and scene-level social reasoning—to quantify the individual and combined impact of the three-step hierarchy.

\subsection{Experimental Setup}
\label{sec:experimental_setup}

\paragraph{Datasets.}
We evaluate our model on three public benchmarks covering synthetic and real-world pedestrian scenarios. 
\textit{JTA}~\cite{ferrari_learning_2018} is a large-scale synthetic dataset with dense multi-agent scenes and 2D/3D pose annotations, used as our primary benchmark. 
\textit{JRDB}~\cite{martin-martin_jrdb_2023,vendrow_jrdb-pose_2023} provides real-world indoor and outdoor trajectories with 2D pose information. 
\textit{Pedestrians and Cyclists in Road Traffic}~\cite{kress_pose_2023} contains real urban pedestrian trajectories with accurate 3D body poses; only the pedestrian subset is used.
Together, these datasets enable evaluation across varying crowd densities, sensing conditions, and pose quality.

\paragraph{Prediction protocol and metrics.}
For JTA and JRDB, models observe 9 frames and predict the next 12 frames at 2.5\,fps. 
For Pedestrians and Cyclists in Road Traffic, we observe 4 frames and predict 12 future frames at 5\,fps due to sequence length constraints.
Training minimizes the Average Displacement Error (ADE), while evaluation reports both ADE and Final Displacement Error (FDE), computed for the pedestrian of interest.

\paragraph{Implementation details.}
All models are trained on a single RTX~6000 Ada GPU using AdamW with a learning rate of $1\times10^{-4}$, weight decay $1\times10^{-2}$, and gradient clipping at 1.0. 
We use a cosine annealing learning rate schedule over 500 epochs and a batch size of 64.
The hierarchical architecture uses a hidden dimension of 192 and a total of six Transformer layers, with four attention heads for each step.
During training, we use stochastic masking to improve robustness to missing observations. Specifically, for each pedestrian and each modality stream (trajectory, 3D pose, bounding boxes, etc.), we randomly drop individual frames with probability 0.1. In addition, with probability 0.3 we drop an entire modality for the whole scene—masking that modality for all pedestrians and all timesteps—so the model learns to operate reliably when a modality is unavailable at inference time.

\subsection{Quantitative Results}
\label{sec:quantitative_results}

\begin{table*}[t]
\centering
\caption{Quantitative results across datasets. ADE and FDE are reported in meters (lower is better).}
\label{tab:all_results}
\setlength{\tabcolsep}{6pt}
\begin{tabular}{l c cc cc cc}
\toprule
\multirow{2}{*}{\textbf{Model}} & 
\multirow{2}{*}{\textbf{Input}} &
\multicolumn{2}{c}{\textbf{JTA}} &
\multicolumn{2}{c}{\textbf{JRDB}} &
\multicolumn{2}{c}{\textbf{Urban}} \\
\cmidrule(lr){3-4} \cmidrule(lr){5-6} \cmidrule(lr){7-8}
& & \textbf{ADE} & \textbf{FDE} & \textbf{ADE} & \textbf{FDE} & \textbf{ADE} & \textbf{FDE} \\
\midrule
Transformer~\cite{giuliari_transformer_2021} & Traj
& 1.56 & 3.54 & 0.56 & 1.10 & 0.60 & 1.11 \\

Social-LSTM~\cite{alahi_social_2016} & Traj
& 1.21 & 2.54 & 0.47 & 0.95 & 0.58 & 1.06 \\

Autobots~\cite{girgis2021latent}  & Traj
& 1.20 & 2.70 & 0.39 & 0.80 & 0.58 & 1.04 \\

EqMotion~\cite{xu_eqmotion_2023} & Traj
& 1.13 & 2.39 & 0.40 & 0.77 & 0.58 & 1.05 \\

ST~\cite{saadatnejad_social-transmotion_2023} & Traj+3D P
& 0.94 & 1.94 & 0.40 & 0.76 & 0.57 & 1.04 \\

EmLoco~\cite{taketsugu_physical_2025} & Traj+3D P
& 0.97 & 1.91 & 0.37 & 0.72 & -- & -- \\

Social-Pose~\cite{gao_social-pose_2025} & Traj+3D P
& \textbf{0.90} & 1.91 & -- & -- & 0.57 & 1.03 \\

\textbf{Ours} & Traj+3D P
& 0.96 & \textbf{1.91} & \textbf{0.33} & \textbf{0.67} & \textbf{0.51} & \textbf{0.96} \\
\bottomrule
\end{tabular}
\end{table*}

We evaluate the proposed hierarchical architecture on the JTA, JRDB, and Pedestrians and Cyclists in Road Traffic datasets using ADE and FDE (lower is better). 
We compare our model with representative baselines, including (i) \emph{trajectory-only} forecasting methods such as Social-LSTM~\cite{alahi_social_2016}, Transformer~\cite{giuliari_transformer_2021}, Autobots~\cite{girgis2021latent} and EqMotion~\cite{xu_eqmotion_2023}, and (ii) \emph{multimodal} or \emph{pose-enhanced} architectures such as Social-TransMotion~\cite{saadatnejad_social-transmotion_2023}, Social-Pose~\cite{gao_social-pose_2025}, and EmLoco~\cite{taketsugu_physical_2025}. All baselines were retrained under the same protocol. This allows us to assess the benefit of multimodal cues and social reasoning independently and jointly.

\paragraph{Overall performance.}
Table~\ref{tab:all_results} summarizes results across datasets. On JTA, multimodal baselines outperform trajectory-based ones, confirming the importance of pose and high-level cues in scenes with many articulated pedestrians. 
Our method is competitive with the best multimodal approaches, achieving an FDE of 1.91~m, and an ADE of 0.96~m.

On JRDB, the improvement is more significant: our approach reduces ADE from 0.37~m (best baselines) to 0.33~m and FDE from 0.72~m to 0.67~m. 
This demonstrates the ability of our model to generalize to challenging real-world indoor and outdoor environments with noisy sensor data.

Finally, on the Urban dataset, trajectory-only methods plateau around 0.57~m ADE and 1.03~m FDE, while multimodal baselines provide only slight gains. 
Our method achieves a substantial improvement, reaching 0.51~m ADE and 0.96~m FDE, showing that our hierarchical fusion is highly effective when high-quality pose cues are available.

\paragraph{Analysis.}
These trends confirm two key insights:  
(i) \emph{multimodal information is crucial} for accurate future motion prediction, especially in scenarios with subtle human intent (turning, slowing, avoiding), and  
(ii) \emph{multi-agent modeling further improves endpoint accuracy}, particularly evident in JTA and JRDB where interactions are frequent.  
The combination of temporal, modality, and scene-level reasoning in our architecture leads to consistent improvements across all benchmarks.

The proposed hierarchical structure achieves consistent gains in both average and final displacement accuracy across synthetic and real datasets, confirming the benefit of structured reasoning across temporal, modal, and social dimensions.

\subsection{Qualitative Results}
\label{sec:qualitative_results}

\begin{figure}[t]
\centering
\includegraphics[width=0.9\linewidth]{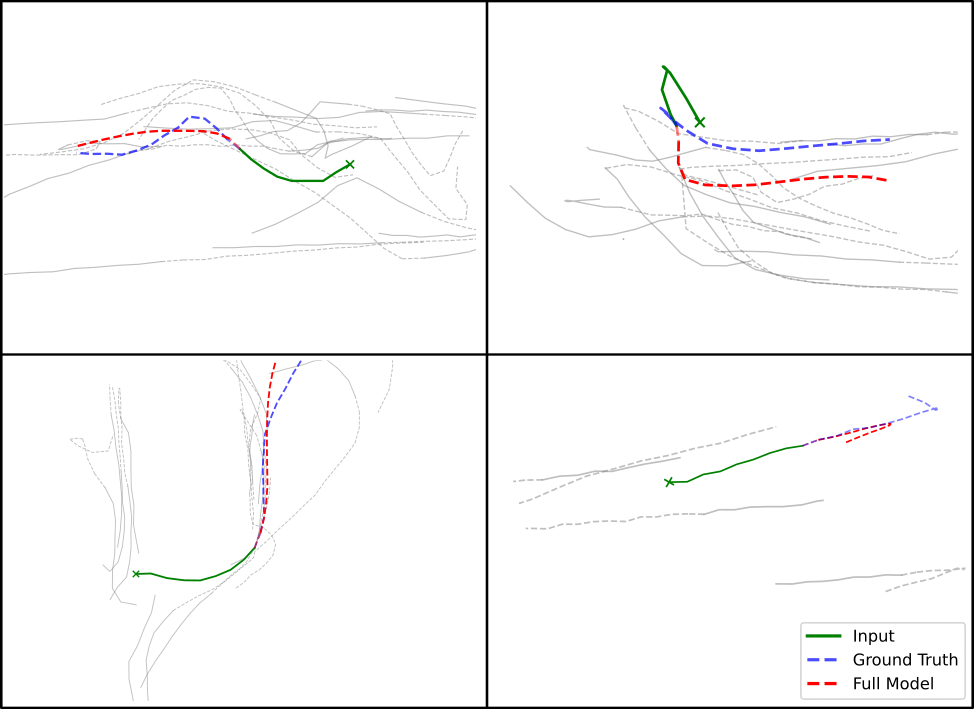}
\caption{Qualitative trajectory prediction examples on JTA scenes.
Observed trajectories are shown in green, ground truth in blue, predictions in red, and other pedestrians in gray.}
\vspace{-0.5cm}
\label{fig:qualitative_results}
\end{figure}

To complement the quantitative evaluation, we illustrate in Fig.~\ref{fig:qualitative_results} several qualitative examples highlighting both the strengths and limitations of our model. In scenes where the future motion is largely determined by recent kinematics (e.g., near-linear progression), the full model closely matches the ground-truth continuation and yields stable predictions with limited drift over the horizon.

In more complex scenes, the model produces forecasts that remain coherent with the overall scene evolution and the motion of nearby pedestrians, but may deviate from the exact future observed in the data. In particular, when several feasible continuations exist (e.g., at bends, local crossings, or when the target pedestrian can either continue or adapt its path), the prediction can follow an alternative yet plausible trajectory. These cases suggest that the social encoder effectively leverages the joint context of all pedestrians---capturing interaction-driven adjustments such as yielding, following, or slight heading changes to avoid conflicts---while the precise continuation taken by the target pedestrian can remain ambiguous from the available observations.

Finally, the bottom-right example illustrates the complementary contribution of pose: even when interactions are limited, the model anticipates an upcoming heading change that is not fully apparent from positions alone, indicating that pose provides early intent cues; however, the turning side can still be misestimated, highlighting remaining uncertainty in fine-grained intent prediction.

\subsection{Ablation Studies}
\label{sec:ablations}

To assess the contribution of each component in our hierarchical architecture, we conduct ablations on four variants that progressively remove multimodal or social information. All variants share the same training and evaluation protocol to ensure comparability.

\begin{table}[h]
\centering
\caption{Ablation of the hierarchical architecture on JTA (lower is better).}
\label{tab:hierarchy_ablation}
\begin{tabular}{lcc}
\toprule
\textbf{Model Variant} & \textbf{ADE} $\downarrow$ & \textbf{FDE} $\downarrow$ \\
\midrule
Single-pedestrian trajectory-only & 1.35 & 2.94 \\
Single-pedestrian multimodal & 1.30 & 2.78 \\
Trajectory-only multi-pedestrian & 1.02 & 2.03 \\
\textbf{Full model (ours)} & \textbf{0.96} & \textbf{1.91} \\
\bottomrule
\end{tabular}
\vspace{-0.5cm}
\end{table}

\paragraph{Architecture hierarchy.}
Table~\ref{tab:hierarchy_ablation} summarizes the performance of the four models.  
The trajectory-only multi-pedestrian baseline already benefits from social interactions.  
Adding multimodal cues without social reasoning (single-pedestrian multimodal) improves temporal anticipation but remains limited by the absence of interaction cues.  
The single-pedestrian trajectory-only model performs the worst, confirming that both multimodality and social context are essential.  
Our full model achieves the best accuracy, demonstrating the complementarity of temporal, modal, and social reasoning.

These results highlight that:  
(i) multimodal cues improve prediction over trajectory-only inputs,  
(ii) social reasoning significantly reduces endpoint error, and  
(iii) combining both yields the best performance.

\begin{figure}[H]
\centering
\includegraphics[width=1.0\linewidth]{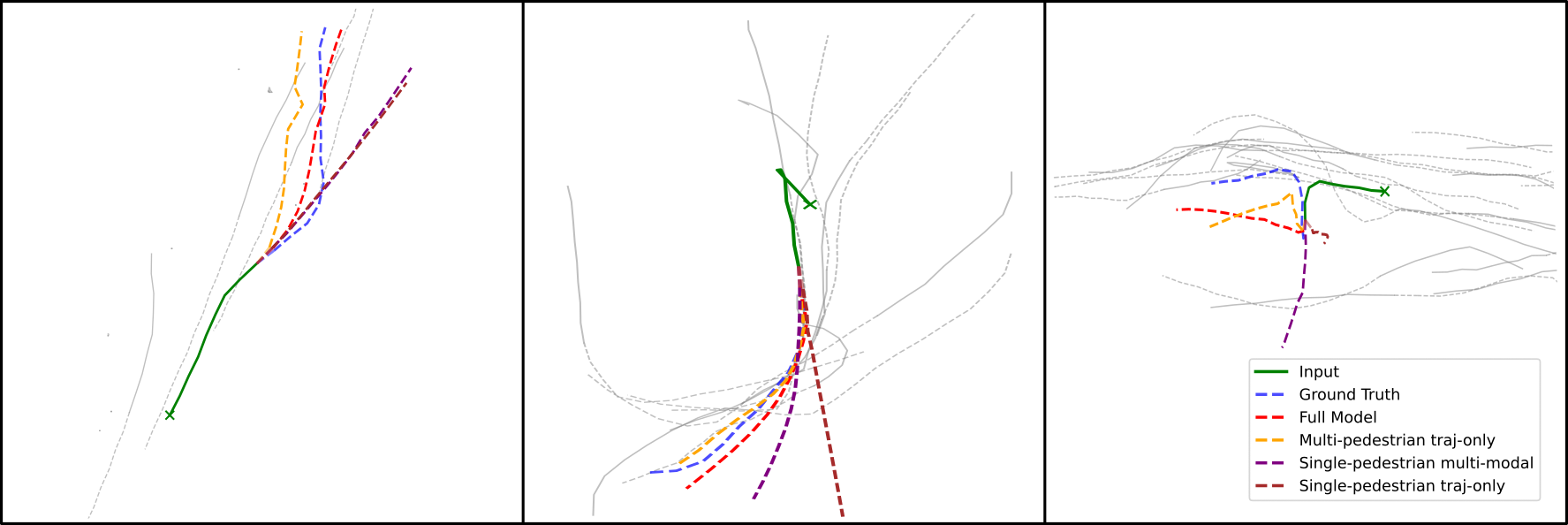} 
\caption{Qualitative comparison of the four hierarchy variants on the same pedestrian and scene. Predictions differ markedly depending on the availability of multimodal cues and social information.}
\label{fig:qualitative_ablation}
\vspace{-0.5cm}
\end{figure}

\paragraph{Qualitative comparison.}
To further illustrate the effect of each component, we visualize predictions for the same pedestrian in the same scene across all four variants.  
Figure~\ref{fig:qualitative_ablation} presents a comparative layout where each variant produces a qualitatively different motion hypothesis, revealing the value of multimodal and social reasoning in avoiding unrealistic drifts and late prediction errors.


\section{Conclusion and future work}\label{sec:conclusion}

We presented a three-step hierarchical Transformer architecture for multi-pedestrian trajectory prediction that decouples temporal encoding, multimodal fusion, and social interaction reasoning. Through lightweight GRU-based modality summarization and scene-level attention over time–agent tokens, the model achieves high computational and memory efficiency, scaling gracefully with the number of modalities and pedestrians while maintaining strong predictive capacity.

Across multiple benchmarks, the proposed architecture demonstrates robust and accurate forecasting, validating the benefits of structured hierarchical reasoning and efficient multimodal integration. Future work will extend this framework toward probabilistic prediction, multi-target forecasting, and richer scene-aware modeling.

\section*{Acknowledgment}
This material is based upon work supported by the ANRT (Association nationale de la recherche et de la technologie) in France with a CIFRE fellowship granted to \href{http://www.explainconsultancy.com/en/}{Explain}.


\bibliographystyle{splncs04}
\bibliography{references}

\end{document}